\documentclass[conference]{IEEEtran}
\IEEEoverridecommandlockouts
\usepackage{cite}
\usepackage{siunitx}
\usepackage{amsmath,amssymb,amsfonts}
\usepackage{algorithmic}
\usepackage{subfig}
\usepackage{graphicx}
\usepackage{textcomp}
\usepackage{xcolor}
\usepackage{hyperref}
\usepackage{tikz}
\usepackage{soul}
\usepackage{booktabs}

\usetikzlibrary{graphs}

\usepackage{adjustbox}

\usepackage{float}
\def\BibTeX{{\rm B\kern-.05em{\sc i\kern-.025em b}\kern-.08em
    T\kern-.1667em\lower.7ex\hbox{E}\kern-.125emX}}

\begin{document}

\title{ROS for Human-Robot Interaction\\

}

\author{\IEEEauthorblockN{Youssef Mohamed}
\IEEEauthorblockA{\textit{University of the West of England} \\
\textit{Bristol Robotics Laboratory}\\
Bristol, UK \\
os19105@bristol.ac.uk}
\and
\IEEEauthorblockN{Séverin Lemaignan}
\IEEEauthorblockA{\textit{University of the West of England} \\
\textit{Bristol Robotics Laboratory}\\
Bristol, UK \\
severin.lemaignan@brl.ac.uk}

}

\maketitle

\begin{abstract}
Integrating real-time, complex social signal processing into robotic systems --
especially in real-world, multi-party interaction situations -- is a
challenge faced by many in the Human-Robot Interaction (HRI) community.  The
difficulty is compounded by the lack of any standard model for human representation
that would facilitate the development and interoperability of
social perception components and pipelines. We introduce in this paper a set of 
conventions and standard interfaces for HRI scenarios, designed to be used with the Robot
Operating System (ROS). It directly aims at promoting interoperability and
re-usability of core functionality between the many HRI-related software
tools, from skeleton tracking, to face recognition, to natural language
processing. Importantly, these interfaces are designed to be relevant to a
broad range of HRI applications, from high-level crowd simulation, to
group-level social interaction modelling, to detailed modelling of human
kinematics. We demonstrate these interface by providing a reference pipeline implementation,
packaged to be easily downloaded and evaluated by the community.

\end{abstract}

\begin{IEEEkeywords}
Human-robot interaction, ROS, Social Signal Processing 
\end{IEEEkeywords}

\section{Introduction}

Social signal processing (both signal detection, and signal interpretation) is a fundamental task in Human-Robot Interaction (HRI). Traditionally, this task is approached through social signal processing
\emph{pipelines}: a combination of software modules, that each
implement a stage of signal processing, and feed their output to the next
module. This pipeline-based approach is common in robotics, for instance for 2D
navigation\footnote{http://wiki.ros.org/navigation}, or 3D image
processing\footnote{http://wiki.ros.org/ecto}. The Robotic Operating System
(ROS)~\cite{quigley2009ros} has played an instrumental roles in enabling quick and iterative design
and implementation of such processing pipelines, by standardizing loosely
coupled data streams (ROS's \emph{topics}) and corresponding datatypes (ROS's
\emph{messages}). And indeed, ROS is today used pervasively in the academic and
industrial robotic communities, as the go-to solution to create real-time data
processing pipelines for complex, real-world sensory information.

Surprisingly, no single effort has been successful creating a similar, broadly
accepted interfaces and pipelines for the HRI domain. As a result, many
different implementations of common tasks (skeleton tracking, face recognition,
speech processing, etc.) cohabit, with their own set of interfaces and
conventions. More concerning for the development of decisional architectures for interactive autonomous robots, the existing software modules are not
designed to work together: a skeleton tracker would typically estimate 3D poses of
bones, without offering any interface for, eg a facial expression recognizer,
to access the face's pixels. A common consequence is that \emph{matching} a 3D
body pose to its corresponding face requires a third-party module, whose role
is to \emph{track} detected skeletons, detected faces (also in case of
temporary occlusions), and associate them. How this 'association' is then
published and shared with the rest of the architecture is effectively
implementation-dependent. Note that we take here the example of matching bodies
to facial expression, but the same could be said of voice processing, speech, gaze estimation, head poses, etc.

The lack of a ROS standard for HRI can be explained both by the relative lack
of maturity of some of the underlying detection and processing algorithms (for
instance, 3D skeleton tracking is a less mature technology than SLAM algorithms used in 2D navigation pipelines), but also by the sheer complexity of HRI
pipelines.
Besides the body/face matching issue mentioned above, we can also mention the highly
variable \emph{scale} (or \emph{granularity}) at which humans are required to be
modeled, depending on the application: from simple, abstract 3D positions in
high-level crowd simulation, to group-level social interaction modelling (that
would for instance require accurate gaze modelling), to accurate modelling of
human kinematics, for eg kinaesthetic teleoperation or Learning for
Demonstration. Also, contrary to most of the objects and situations traditionally encountered in robotics, humans are bodies that are typically not known prior to runtime, and are highly dynamic: it is commonly expected
that they will appear and disappear from the robot sensory space multiple times
during a typical interaction. This transient nature causes various issues,
including a need for robust tracking, re-identification, managing a history of
known people, etc.

In order to provide robust, complete foundations on which to address these
issues, we present in this article the \emph{ROS4HRI} framework, aiming at:

\begin{itemize}
    \item Identifying, designing and implementing an appropriate, ROS-based
        \emph{representation system} for humans, both appropriate for a 
        broad range of HRI applications (from a single individual to crowds),
        and practical with respect to available tools for social signal
        processing;

    \item The specification of a reference processing pipeline, that effectively
        implement a modular, loosely-coupled framework for social signal
        processing, able to integrate multiple modalities when available, and
        scalable from a single user to large groups.

\end{itemize}

Alongside these two specifications, we also present an open-source
implementation of the ROS4HRI framework, that currently covers a subset of the
specifications, namely the 3D tracking and matching of skeletons and faces in
groups of up to about 10 people. The main open-source code repository can be found here:
\href{https://github.com/ros4hri/ros4hri}{\tt github.com/ros4hri/ros4hri}.

Achieving these goals will allow much better collaboration between projects and
allow a reduction in duplicate efforts to implement the same functionality.

The remaining of the article is structured in the following way: we review next
previous work pertaining to ROS and HRI; we then introduce our human model (made
of four components: the body, the face, the voice and the person); we present
next the ROS implementation of our model, a combination of a limited set of new,
HRI ROS messages, and a particular topic structure; we then present a
specification of the HRI pipeline, and finally introduce a reference
implementation, validated on a small dataset of naturalistic social
interactions.

\section{Related work}

Social signal processing in robotics is a broad topic, and we do not review here
specific algorithms (we can refer the interested reader
to~\cite{burgoon2017social} as an introduction to social signal processing, and
to the numerous surveys already published on specific social signals processing
techniques).

We look hereafter first into some significant \emph{non-ROS} social signal
processing approaches, then we cover the (limited) early attempts at creating
ROS interfaces for HRI, and finally, we discuss a few ad-hoc projects which used
ROS for HRI, without attempting to build a generic, application-agnostic
framework out of it.

\subsection{Approaches to social signal processing in HRI}

Several frameworks have been developed over the years for HRI;
for example,~\cite{fong2006human} introduced the human-robot interaction
operating system (HRI/OS). HRI/OS is an architecture that allows cooperation
between humans and robots. The HRI/OS supported peer-to-peer dialogue, and the
architecture introduced a way to assign tasks to the agents. The agent is able
to ask for help if needed from the human, based on the information programmed
into the robot about the human. HRI/OS lacked a higher level of autonomy, as it
does not collect information about humans. Nonetheless, it introduced the idea
of creating a framework for HRI. 

The LAAS architecture for social autonomy~\cite{lemaignan2017artificial} is
another framework featuring real-time modelling of human interactors. SHARY,
their architecture controller, aimed to enhance the collaboration between humans
and robots by introducing a layered architecture for decision planning.
Nonetheless, the framework also considered the human's position and gaze
direction, which had a direct effect on the decision planning process that the
robot had to compute.

All the discussed frameworks focused mainly on developing the decision planning
architecture with little focus on the human's social signals (i.e. body
language, emotional speech, facial expressions) and underlying behaviour.
Therefore, The Social Signal Interpretation (SSI) framework
\cite{wagner2013social} has introduced an approach, that social signals can be
recorded, analyzed and classified in real-time. The patch-based design of the
SSI allows numerous types of sensors to be integrated with the ability for all
of them to work in parallel and synchronize the input signals. Furthermore, SSI
supports the use of machine learning models, as it has a graphical user
interface which aids in the process of annotating the data and then integrating
the models created in the data extraction process. 

\subsection{ROS and HRI}


Only a few attempts have been made in the literature to utilize ROS as a social
signal extraction method, often focusing in one type
of social signals, ignoring the others due to their complexity. 

To the best of our knowledge, only two ROS projects have attempted to create a
stand-alone toolset for HRI: the \href{https://github.com/wg-perception/people}{\tt
people}\footnote{\url{https://wiki.ros.org/people}} package, originally
developed by Pantofaru in 2010-2012 (last code commit in 2015), and the
\href{https://github.com/ipa320/cob_people_perception}{\tt
cob\_people\_perception}\footnote{\url{http://wiki.ros.org/cob_people_detection}}
package~\cite{bormann2013person}, developed in 2012-2014 in the frame of the EU
project ACCOMPANY (and still maintained).

Neither of these two attempts is however generic in the sense that they propose
a complete, multi-modality, technology-agnostic approach: the {\tt people}
package had a narrow scope (leg tracking and face tracking), and the {\tt
cob\_people\_perception} stack is mainly built around the Kinect hardware and
NITE software library. However, some of the HRI ROS messages we introduce hereafter have
roots in these two early attempts.

On the matter of representing the human body using ROS conventions, we draw our
naming conventions from the work done in humanoid robots. Specifically, the ROS
REP-120\footnote{\url{https://www.ros.org/reps/rep-0120.html}} partially defines
a naming convention for humanoid robots that we follow here to a large extend.

The Human-Robot Interaction toolkit~\cite{lane-etal-2012-hritk} ({\tt HRItk}) is
another ROS package for speech processing. This is done by integrating several
natural language modules, like speech detection and recognition, natural
language understanding and dialogue state analysis. {\tt HRItk} also has two
basic models for gesture recognition and gaze tracking, both of which were basic
concepts and are not maintained in the toolkit.  Nonetheless, the toolkit
provided an efficient architecture for NLP using ROS, and was the bases of other
architectures in the literature~\cite{zhang2015ros}\cite{li2013towards}. On the
other hand, it does not cover the uses of other social signals, like body
language and facial expressions.

\subsection{Ad-hoc ROS-based pipelines for HRI}

There are several projects that are being discussed in the literature that are
trying to achieve an integration between the spatio-tempo awareness of a robot
and the social understanding of social situations. For example, STRANDS has been
covering a range of issues in the HRI field, from world mapping to human
activity recognition. Nonetheless, in their
paper~\cite{10.5555/2906831.2906838}, attempt has been made to integrate a robot
in physical therapy sessions for older adults with dementia. The approach was
successful and the robot was able to have some positive effects on the patients
while being partially controlled by the therapist through using cards with
instructions for the robot. Nonetheless, the approach concluded that, a better
understanding of the patients was needed when they were trying to interact with
the robot, as most of them found it hard to use a touch screen for
communication. Hence, the use of better understanding of group dynamics and the
relationship between the therapist and the patient would make the interaction
significantly easier than the robot being highly dependant on the cards shown to
it by the therapist. Similarly, the POETICON++ project aims to achieve similar
aims and covers several aspect of HRI, but mainly focusing on natural language
processing \cite{Twomey_Morse_Cangelosi_Horst_2016,badino2016integrating}.
However, several publications also focused on discussing cognitive abilities for
social interactions~\cite{book,michael2015domain}. The project shows the
significance of having an understanding of social situations and the uses for
such cognitive abilities in HRI.

\section{The ROS4HRI human model}

\subsection{The four human identifiers}

\begin{figure}
    \centering
    \includegraphics[width=\linewidth]{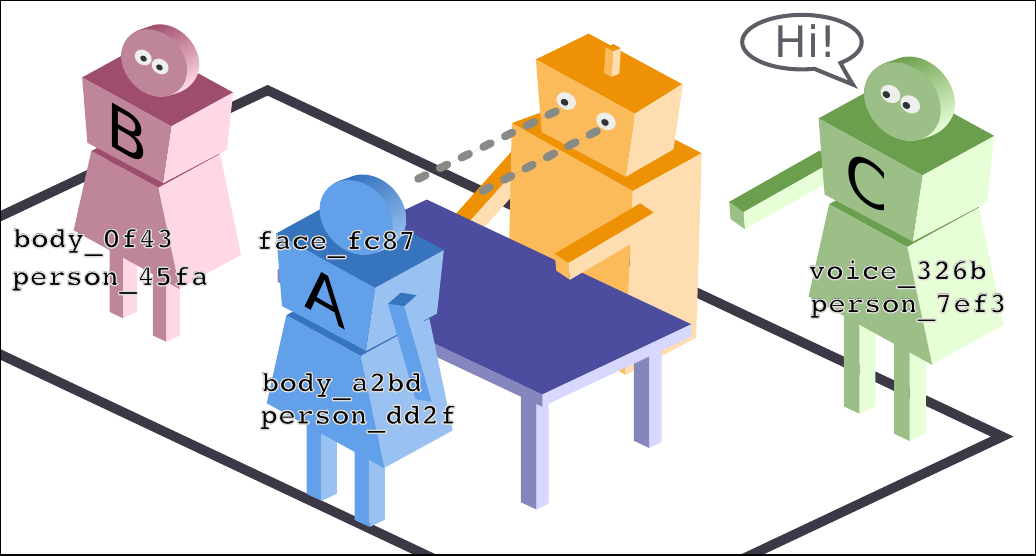}
    \caption{In this situation: A is facing the robot: A gets a unique {\tt
    faceID}, a unique {\tt bodyID}, and a unique {\tt personID}; B's body
    is visible to the robot, but not the face: B only gets a {\tt bodyID} and
    {\tt personID}; C is not seen, but heard: C gets a {\tt voiceID} and a
    {\tt personID}.}
    \label{fig|social-situation}
\end{figure}

To accommodate existing tools and technique used to detect and recognize humans,
the representation of a person is built on a combination of 4 unique
identifiers: a face identifier, a body identifier, a voice identifier and a
person identifier. These four identifiers are not mutually exclusive, and
depending on the requirements of the application, the available sensing
capabilities, and the position/behaviour of the humans, only some might be
available for a given person, at a given time (see
Figure~\ref{fig|social-situation}).

\paragraph{Face identifier}

The face identifier is a unique ID (UUID) that identifies a detected face.
This UUID is typically generated by the face detector/head pose estimator upon
face detection.  There is a one-to-one relationship between this face ID and the
estimated 6D pose of the head, represented as a ROS TF frame named
\texttt{face\_<faceID>} (cf below for details regarding the face frame conventions).
Importantly, this ID is \textbf{not persistent}: once a face is lost (for
instance, the person goes out of frame), its ID is not valid nor meaningful any
more. In particular, there is no expectation that the face detector will
attempt to recognise the face, to re-assign the same face ID if the person
re-appears.

At any given time, the list of tracked faces is published under the {\tt
humans/faces/tracked} topic.

\paragraph{Body identifier}

The body identifier is similar to the face ID, but for a person's skeleton. It
is typically created by the skeleton tracker upon detection of a skeleton.  Like
the face ID, the body ID is \textbf{not persistent} and is valid only as long as the
specific skeleton is tracked by skeleton tracker which initially detected it.
The corresponding TF frame is \texttt{body\_<bodyID>}, and TF frames associated
to each of the body parts of the person are suffixed with the same ID (cf
below).

The list of tracked skeletons is published under the {\tt humans/bodies/tracked} topic.

\paragraph{Voice identifier}

Likewise, a voice separation module should assign a unique, non-persistent, ID
for each detected voice.

The list of tracked skeletons is published under the {\tt humans/voices/tracked} topic.

\paragraph{Person identifier}

Finally, the person identifier is a unique ID \textbf{permanently} associated with a
unique person. This agent ID should assigned by a module able to perform person
identification (face recognition module, voice recognition module, sound source
localization + name, identification based on physical features like
height/age/gender, person identification based on pre-defined features like the
colour of the clothes, etc.) This ID is meant to be persistent so that the robot
can recognize people across encounters/sessions.

When meaningful, a TF frame must be associated with the agent ID, named
\texttt{person\_<personID>}. Due to the importance of the head in human-robot
interaction, the \texttt{person\_<personID>} frame is expected to be placed as
close as possible to the head of the human.  If neither the face nor the
skeleton is tracked, the \texttt{person\_<personID>} frame might be located to
the last known position of the human or removed altogether if no meaningful
estimate of the human location is available. We detail below the rules
associated to the \texttt{person\_<personID>} frame.

\paragraph{Interactions between the different identifiers}

Table~\ref{table|ids} presents examples of the various possible combination of
identifiers, with the corresponding interpretation.

\begin{table*}[ht]
    \centering
        \caption{Interpretation of different identifiers combinations (note that, for brevity, not all possible combinations are presented)}
        \label{table|ids}
    \resizebox{\textwidth}{!}{
        \begin{tabular}{@{}llllp{11cm}@{}}
\toprule
\bf Face ID & \bf Body ID & \bf Voice ID & \bf Person ID & \bf Interpretation                                                                                                                                                                                                                                                                                \\ \midrule
{\tt 24ac}  & Ø           & Ø            & Ø             & Face detected - random id {\tt 24ac} assigned - corresponding TF frame {\tt face\_24ac} is published.                                                                                                                                                                                             \\\midrule

{\tt d73b}  & Ø           & Ø            & Ø             & Face detected (possibly a re-detection of a previous one) - random id {\tt d73b} assigned + frame published.                                                                                                                                                                                      \\\midrule

Ø           & {\tt 37ef}  & Ø            & Ø             & Skeleton detected - id {\tt 37ef} assigned + frame {\tt body\_37ef} published.                                                                                                                                                                                                                    \\\midrule

{\tt d73b}  & {\tt 37ef}  & Ø            & Ø             & A face/body matcher associated the face and the skeleton together.                                                                                                                                                                                                                                \\\midrule

Ø           & Ø           & Ø            & {\tt 9d8a}    & Person {\tt 9d8a} is known, but not associated with any face, body or voice. Note that TF frame {\tt person\_9d8a} might nevertheless exist (for instance, the last known position of the human).                                                                                                 \\\midrule

{\tt d73b}  & Ø           & Ø            & {\tt 9d8a}    & Face {\tt d73b} is associated to person {\tt 9d8a}. Typical result of successful face recognition.                                                                                                                                                                                                \\\midrule

{\tt 96f1}  & Ø           & Ø            & {\tt 9d8a}    & Person {\tt 9d8a} is now associated to face {\tt 96f1}: this new association might come from the face tracker losing track of a previous face, thus re-assigning a different id to the face. The newly assigned face is however recognized by the face recognition module as being {\tt person\_9d8a}. \\\midrule

{\tt 96f1}  & {\tt 37ef}  & Ø            & {\tt 9d8a}    & The human {\tt 9d8a} is fully tracked: both the head and the body are detected.                                                                                                                                                                                                                   \\\midrule

Ø           & Ø           & {\tt ab7f}   & {\tt baf0}    & A voice has been isolated, and assigned to a new (likely unseen) person {\tt baf0}                                                                                                                                                                                                                \\\midrule

Ø           & Ø           & Ø            & Ø             & This is not permitted: at least one identifier must exist.                                                                                                                                                                                                                                        \\ \bottomrule
\end{tabular}

    }
\end{table*}

\section{ROS implementation}

\subsection{Topics structure}

Our implementation exposes social signals using a specific structure of ROS
topics, and introduce a limited number of new ROS messages.

We propose the following rules to present human perceptions in a ROS system:

\begin{enumerate}
    \item all topics are grouped under the global namespace \texttt{/humans}
    \item five sub-namespaces are available:
        \begin{itemize}
            \item \texttt{/humans/faces}
            \item \texttt{/humans/bodies}
            \item \texttt{/humans/voices}
            \item \texttt{/humans/persons}
            \item \texttt{/humans/interactions}
        \end{itemize}
    \item the first four (\texttt{/faces}, \texttt{/bodies}, \texttt{/voices},
        \texttt{/persons}) expose one sub-namespace per face, body, voice, person
        detected, named after the corresponding id: for instance,
        \texttt{/humans/faces/<faceID>/}. In addition, they expose a topic
        {\tt /tracked} where the list of currently tracked faces/bodies/voices/persons is
        published.

\end{enumerate}

The structure of each sub-namespace is presented in Table~\ref{table|topics}.

\begin{table*}[ht!]
    \caption{Topic structure for human-related signals (the newly introduced
    {\tt hri\_msgs} message types are visible in the \emph{Message type}
    column)}
\label{table|topics}
    \begin{tabular}{p{3cm}p{5cm}p{8cm}}
\toprule
        \multicolumn{3}{l}{{\tt /humans/faces/<faceID>} (for instance, {\tt /humans/faces/bf3d})}  \\
\toprule
\textbf{Name} & \textbf{Message type}             & \textbf{Description}                                                \\ \midrule
\tt /roi          & \tt sensor\_msgs/RegionOfInterest & Region of the face in the source image                              \\
\tt /landmarks    & \tt hri\_msgs/FacialLandmarks     & The 2D facial landmarks extracted from the face                     \\
\tt /facs         & \tt hri\_msgs/FacialActionUnits   & The presence and intensity of facial action units found in the face \\
\tt /expression   & \tt hri\_msgs/Expression          & The expression recognised from the face                             \\ \bottomrule
\multicolumn{3}{l}{}                                                 \\
\multicolumn{3}{l}{\tt /humans/bodies/<bodyID>}                                                 \\
\toprule
\textbf{Name} & \textbf{Message type}             & \textbf{Description}                                                \\ \midrule
\tt /roi          & \tt sensor\_msgs/RegionOfInterest & Region of the whole body in the source image                              \\
\tt /skeleton2d    & \tt hri\_msgs/Skeleton2D       & The 2D points of the detected skeleton                     \\
\tt /attitude         & \tt hri\_msgs/BodyAttitude   & Recognised body attitude or gesture \\ \bottomrule
        \multicolumn{3}{l}{\emph{(see below for 3D skeletons and poses, which are represented through TF frames)}}                                                 \\
\multicolumn{3}{l}{}                                                 \\
\multicolumn{3}{l}{\tt /humans/voices/<voiceID>}                                                 \\
\toprule
\textbf{Name} & \textbf{Message type}             & \textbf{Description}                                                \\ \midrule
        \tt /audio    & \tt audio\_msgs/AudioData & Separated audio stream for this voice \\
        \tt /features    & \tt hri\_msgs/AudioFeatures & INTERSPEECH'09 Emotion challenge~\cite{schuller2009interspeech} low-level audio features. \\
        \tt /is\_speaking    & \tt std\_msgs/Bool & Whether or not speech is recognised from this voice            \\
        \tt /speech    & \tt std\_msgs/String       & The live stream of speech recognized via an ASR engine \\ \bottomrule
\multicolumn{3}{l}{}                                                 \\
        \multicolumn{3}{l}{{\tt /humans/persons/<personID>}}                                                 \\
\toprule
\textbf{Name} & \textbf{Message type}             & \textbf{Description}                                                \\ \midrule
\tt /face\_id          & {\tt std\_msgs/String} (latched) & Face matched to that person (if any)                              \\
\tt /body\_id          & {\tt std\_msgs/String} (latched) & Body matched to that person (if any)                              \\
\tt /voice\_id          & {\tt std\_msgs/String} (latched) & Voice matched to that person (if any)                              \\
\tt /location\_confidence & {\tt std\_msgs/Float32} & Location confidence; 1 means 'person current seen', 0 means 'person location unknown' \\
\tt /demographics    & \tt hri\_msgs/AgeAndGender       & Detected age and gender of the person                     \\
\tt /name         & \tt std\_msgs/String   & Name, if known\\
\tt /native\_language         & \tt std\_msgs/String   & IETF language codes like \texttt{EN\_gb}, if known\\ \bottomrule
\multicolumn{3}{l}{}                                                 \\
\multicolumn{3}{l}{\tt /humans/interactions}                                                 \\
\toprule
\textbf{Name} & \textbf{Message type}             & \textbf{Description}                                                \\ \midrule
\tt /groups          & \tt hri\_msgs/GroupsStamped & Estimated social groups                              \\
\tt /gaze            & \tt hri\_msgs/GazesStamped & Estimated gazing behaviours                              \\ \bottomrule
\end{tabular}
\end{table*}

\subsection{The \texttt{hri\_msgs} ROS messages}

Table~\ref{table|msgs} lists the newly introduced ROS messages for HRI. They are
regrouped in the {\tt hri\_msgs} ROS package.

\begin{table}[ht]
\caption{List of newly introduced ROS messages for HRI}
\label{table|msgs}
    \begin{tabular}{p{2.5cm}p{5.6cm}}
\toprule
\textbf{Message name} & \textbf{Motivation}
    \\ \midrule
    {\tt AgeAndGender}          & As mentioned in~\cite{kuo2009age}, age and
                                  gender are key demongraphic factors when it
                                  comes to user acceptance of robots. The
                                  message encode both age and gender, with
                                  associated levels of confidence.
                                  \\ \midrule

    {\tt AudioFeatures}          & Encodes 16 low-level audio features, based on
                                   the INTERSPEECH'09 Emotion recognition
                                   challenge~\cite{schuller2009interspeech}.\\ \midrule

    {\tt BodyAttitude}          & Body posture recognition is essential when
                                  designing cooperative
                                  robots~\cite{gaschler2012social}. The message
                                  encodes three such categorical body postures (hands on
                                  face, arms crossed, hands raised), and could
                                  be easily extended in the future. \\
                                  \midrule

    {\tt Expression}            & Expressions and basic emotions are extensively
                                  discussed in the literature due to the amount of information they infer
                                  about human behaviour. The {\tt Expression}
                                  message encode facial expression, either in a
                                  categorical manner (Ekman's
                                  model~\cite{ekman1992argument}), or using the
                                  Valence/Arousal continuous plane.
                                  \\ \midrule

    {\tt FacialAction Units}     & Encodes the intensity and confidence level of
                                  detected Facial Action Units, following the
                                  coding scheme and nomenclature proposed
                                  in~\cite{ekman1978facs}.
                                  \\ \midrule

    {\tt FacialLandmarks}       & Encodes the 2D coordinates in image space (and
                                  confidence) of 67 facial landmarks (including
                                  mouth, nose, eyes, and face silhouette).
                                  \\ \midrule

    {\tt Group}                 & List of person IDs being detected as forming a
                                  social group. The list of all groups is
                                  published as a {\tt GroupsStamped} message.    \\ \midrule

    {\tt GazeSender Receiver}    & Encodes one person being observed as gazing at another, as a
                                  pair of person IDs. The list of all such
                                  gazing behaviour at a given time is published
                                  as a {\tt GazesStamped} message. \\ \midrule

    {\tt Skeleton2D}            & The message encodes the 3D coordinates of 18
                                  skeletal key points.
    \\ \bottomrule

\end{tabular}
\end{table}

\subsection{Human kinematic model}

\begin{figure}
    \centering
    \includegraphics[width=\linewidth]{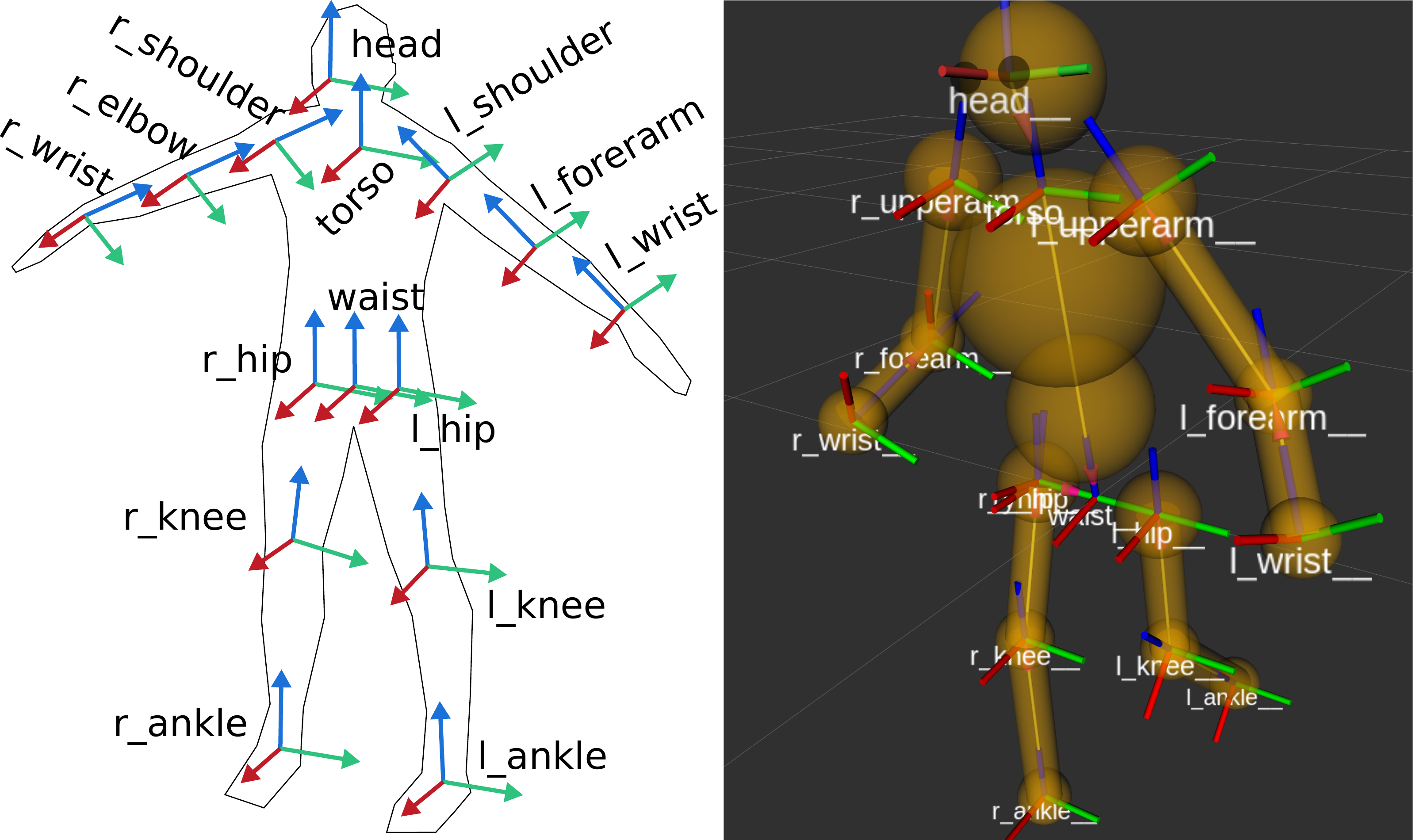}
    \caption{Left: the 15 links defined on the human body. Frames orientations and
    naming are based on REP-103 and REP-120. Right: the URDF kinematic model, viewed in RViz.}
    \label{fig|frames}
\end{figure}

URDF\footnote{\url{http://wiki.ros.org/urdf}} is the XML-based language used by
ROS to represent kinematic models. Besides visualization, URDF models are used
by several ROS tools to reason about the kinematic of systems (for instance, for
motion planning or grasp planning). In order to leverage these tools, we adopt a
URDF-centric approach to human kinematics.

However, unlike robots (whose kinematic models are usually fixed, and known
beforehand), humans anatomies do vary, and in principle, each person would
require a unique kinematic model, reflecting her/his individual height, weight, etc.

We model individual difference by generating on-the-fly custom URDF models every
time a person is detected, using the person's observed height as the main
parameter, from which other dimensions (like the shoulder-to-shoulder width, the
waist width, and the length of the limbs) are derived, based on standard models of
anthropometry.

The generated URDF model is then published on the ROS parameter server (using
the {\tt bodyID} unique identifier), making it available to the rest of the ROS
network.

The URDF model is used in combination with the computed joint state of each
tracked body to then generate a kinematically-sound, real-time 3D model of the
person (Figure~\ref{fig|frames}).

\subsection{Frame conventions}

The ROS4HRI specifies several TF frames to spatially represent a human.

Where meaningful, the HRI frames follow the conventions set out in the
\emph{REP-103 -- Standard Units of Measure and Coordinate
Conventions}\footnote{\url{https://www.ros.org/reps/rep-0103.html}}, and where
relevant, \emph{REP-120 -- Coordinate Frames for Humanoid
Robots}\footnote{\url{https://www.ros.org/reps/rep-0120.html}}.

\subsubsection{Body frames}


Figure~\ref{fig|frames} shows the 15 frames defined on the human skeleton. The
{\tt waist\_<bodyID>} is collocated with the body's root frame,
\texttt{body\_<bodyID>} (where \texttt{<bodyID>} stands for the unique body
identifier). The origin of this frame is located at the midpoint between the two
hips, and the parent of this frame would typically be the sensor frame used to
estimate the body pose. All skeleton points published as TF frames are suffixed with the same
\texttt{<bodyID>}, thus enabling several unique skeletons to be tracked and
visible in TF simultaneously (not visible on Fig.~\ref{fig|frames} for clarity).

Following the REP-103, the $x$-axis of the frames points forward (i.e., out of the body),
while the $z$-axis points toward the head (i.e. up when the person is standing
vertically, with arm resting along the body).

The 15 links are connected through 18 joints: 3 degrees of freedom (DoF) for the
head, 3 DoFs for each shoulder, 1 DoF for elbows and knees, 2
DoFs for the hips, and 1 DoF for the waist. In the current version, the wrists
and ankles are not articulated (due to the lack of support for tracking hands
and feet in 3D pose estimators), but this could be easily added in future
revisions.

\subsubsection{Face frame}

Head pose estimation modules are requested to publish the 6D head pose as a TF frame
named \texttt{face\_<faceID>} where \texttt{<faceID>} stands for the unique face
identifier of this face.

The parent of this frame is the sensor frame used to estimate the face pose.
The origin of the frame must be the sellion (defined as the deepest midline
point of the angle formed between the nose and forehead. It can generally be
approximated to the midpoint of the line connecting the two eyes).

The $x$-axis is expected to point forward (i.e., out of the face), the $z$-axis is
expected to point toward the scalp (i.e., up when the person is standing
vertically).

\textbf{Head vs face frames} If the skeleton tracker provides an estimate of the head pose, it might publish
a frame named \texttt{head\_<bodyID>}, located at the sellion (mid-point between
the two eyes). It is the joint responsibility of the face
tracker and skeleton tracker to ensure that \texttt{face\_<faceID>}
\texttt{head\_<bodyID>} are consistent with each other, e.g. collocated.

\textbf{Gaze} In addition to the face, a head pose estimator might publish a TF frame
representing the gaze direction, \texttt{gaze\_<faceID>}. The {\tt gaze} frame
is normally collocated with the {\tt face} frame. However, it  follows the
convention of cameras' optical frames: the $z$-axis points forward, the $y$-axis
points down.

\subsubsection{Person frame}

The \texttt{person\_<personID>} frame has a slightly more complex semantic and
needs to be interpreted in conjunction with the value published on the topic {\\
/humans/persons/<personID>/location\_confidence}.

We can distinguish four cases:

\begin{itemize}
    \item The person has not yet been identified; no {\tt personID} has been assigned
        yet. In that case, no TF frame is published. In other words, the TF
        frame \texttt{person\_<personID>} can only exist once the person has been
        identified (and, as such, can be later re-identified).

    \item The human is currently being tracked (i.e. {\tt personID} is set, and at
        least one of {\tt faceID}, {\tt bodyID} or {\tt voiceID} is set). In
        this case, {\tt location\_confidence} should be 1, and:

        \begin{enumerate}
            \item if a face is associated to the person, the \texttt{person\_<personID>} frame
                must be collocated with the \texttt{face\_<faceID>} frame.

            \item else, if a body is associated to the person, the
                \texttt{person\_<personID>} frame must be collocated with the
                skeleton frame the closest to the head.
            \item else, the best available approximation of the person's position (for
                instance, based on sound source localization) should be used.

        \end{enumerate}

    \item The human is not currently seen/heard, but a prior localization is
        known. In this case,
        \texttt{location\_confidence} must be set to a value $<$ 1 and a person
        \texttt{person\_<personID>} TF frame must be published as long as
        \texttt{location\_confidence} $>$ 0. Simple implementations might choose
        to publish \texttt{location\_confidence} $=$ 0.5 as soon as the person is not
        actively seen anymore, while continuously broadcasting the last known location.
        More advanced implementations might slowly decrease
        \texttt{location\_confidence} over time (to represent the fact that the human might
        have walked away, for instance), eventually stopping to publish the
        {\tt person\_<personID} frame.

    \item The system knows about the person (for instance, from dialogue with
        another person), but has no location information. In this case,
        \texttt{location\_confidence} must be set to 0, and no TF frame
        should be broadcast.

\end{itemize}

\section{Reference pipeline}

\subsection{Generic pipeline specification}

\begin{figure*}
    \centering
    \includegraphics[width=0.9\linewidth]{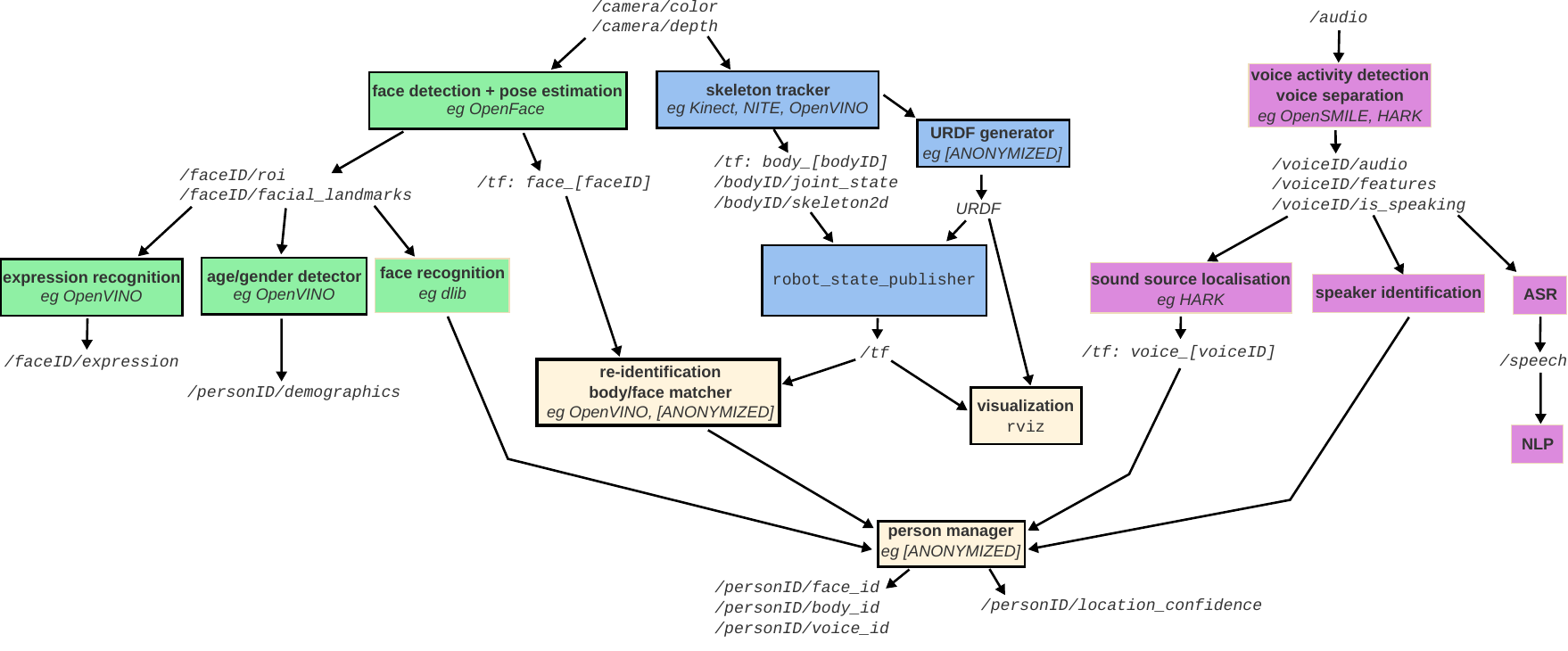}
    \caption{Reference signal processing pipeline. Green nodes (left) process
    facial signals, blue nodes (middle) deal with the body tracking, while
    purple nodes (right) implement the audio processing part. Light yellow
    nodes, at the bottom, deal with modalities fusion, and manage the permanent
    {\tt personIDs}. Only nodes with a strong border are present in our
    reference pipeline implementation.}
    \label{fig|pipeline}
\end{figure*}

So far, we have presented a generic model for human representation, targeted to
HRI, and we have shown how this model could be implemented using ROS conventions
and tools.

This section presents a reference pipeline that could be implemented and
deployed to acquire and process social signals in real-time, making use of the
proposed model. We then present a partial implementation of this generic
reference pipeline, that focuses on face and bodies.

Figure~\ref{fig|pipeline} represents our reference pipeline. Importantly, this
is \emph{not} a normative reference: its purpose is rather to illustrate how a
set of ROS nodes can be organized into a social signal processing framework
which makes full use of the ROS4HRI models. In particular, the node
implementations are not specified (even though we suggest implementations for
some of them).

Also, the split between nodes as pictured in Figure~\ref{fig|pipeline} is
somewhat arbitrary: depending on a given implementation, several functionalities
might be offered together or not: while a module like OpenFace~\cite{inproceedings}
provides face detection, head pose estimation, facial landmark detection, and
facial action units extraction in one package, alternatives are possible, using
for instance {\tt dlib}~\cite{king2009dlib} for landmark detection and face recognition,
and {\tt gazr}~\cite{lemaignan2016realtime} for head pose estimation.

The possibility of flexibly shaping the processing pipeline is a key aspect of
the ROS4HRI project, making it possible to tailor the pipeline to the need of
the target application, or to the availability (or not) of specific sensors and
compute capabilities. By relying on well-defined interfaces, the ROS4HRI project
also enables modular design, where one can iteratively improve (or replace for
better implementations) some parts of the system without impacting the rest.

\subsection{Reference implementation}



Our reference pipeline extracts and represents the following features:

\begin{itemize}

    \item \textbf{Facial landmarks}: facial landmarks are used to determine the
        action units and can be useful in so many other ways depending on the
        application needed, hence, OpenFace is used to detect the facial
        landmarks. 

    \item \textbf{Action units}: as action units depend on the landmarks
        detected, they can infer the emotions of the person and has multiple
        other uses. 

    \item \textbf{Face 3D position}: The face position in real-life units can be
        the most crucial feature, as it utilizes the head size to estimate the
        3D position without the depth information. This can infer proximity
        between people and gaze direction. 

    \item \textbf{Gaze direction}: The gaze is detected using the transformed
    frames produced by the 3D position of each of the heads and can detect which
head is looking at the other.

    \item \textbf{Age and Gender}: OpenVino is used
    to detect the age and gender.

    \item \textbf{2D and 3D skeletal key-points}: 18 body key-points are detected using
        OpenVINO, both in 2D and in 3D, and also supporting multiple people. The
        3D keypoints are used to generate on-the-fly URDF models of the detected
        persons, as well as computing their joint state (using the {\tt ikpy} inverse
        kinematics library\footnote{\url{https://github.com/Phylliade/ikpy}}).
        Automatically-spawned instances of ROS's {\tt robot\_state\_publisher}
        are then responsible for publishing a kinematically consistent TF tree
        for each person.
       
    \item \textbf{Body pose}: the upper body pose is detected by using the
        distances between the first 7 points detected by the OpenPose COCO model
        and can classify: hands-on face, hands raised and arms crossed. All
        three classifications can infer the degree of engagement of the person
        in the interaction. 
    \end{itemize}

\subsection{Evaluation}

To be able to evaluate the pipeline created, an environment that is rich in social signals had to be created. Therefore, a data set of 3 people playing the deception based role-playing game, mafia, has been recorded. The reference pipeline was able to extract the features discussed. Nonetheless, pipeline showed significant CPU contention as several models were running in parallel. The models ran in 4 FPS, on an \emph{Intel Core i7-6700HQ CPU @
2.60GHz} CPU. Only CPU was used as most models did not have GPU support. Furthermore, the gaze direction algorithm has been evaluated by comparing the annotated gaze of the players with the detected gaze instances by the algorithm. The gaze detection algorithm was able to detect 78\% of the gaze instances. 

\section{Discussion and future work}



\subsection{Integration into the ROS ecosystem}

We aim at submitting a \emph{ROS Enhancement Proposal} (REP) to formally specify
the ROS4HRI proposal once a 'sufficiently large' amount of HRI practitioners will have
read the proposal, and provided feedback. As such, this article also aims at
engaging the community with this design effort. We will use the project's public
issue tracker to record the feedback, and further discuss and refine the
proposal with the community.

In terms of ROS integration, we have decided to elect ROS1 instead of ROS2,
mostly due to the familiarity of the authors with ROS1, and the extensive
amount of code and algorithms available within the ROS1 ecosystem. Once the
ROS4HRI design is fully stabilised (eg, after engaging with the community), we will
certainly consider porting it to ROS2. In particular, the messages and topics
structure should be straightforwardly transferable.

\subsection{Reference pipeline}

As some of the social signals are dependant on each other, some relationships
have already been made in the pipeline. Nonetheless, a connection between the
gaze detection and the action units can be useful in the future. Action units
can detect the movements of specific muscles in the face, and action unit 45 is
associated with blinking. Hence, integrating the action of blinking (or eyes
closed) can lead to better detections in the gaze algorithm. As when the eyes
are closed the algorithm in its current state would still falsely detect that
the person is looking to the other person in the direction of gaze. In the case
of which the action units are integrated, the detection would only be made if
the person has there eyes open. Adding this feature to the system would increase
the accuracy significantly, especially in the case of playing MAFIA, as the
participants are required to close their eyes during the night phase of the
game. 

In addition, CPU contention was one of the main issues that have been faced
during the testing of the system, hence, making the toolkits used compatible
with the machine's GPU would increase the performance significantly. Also, it
would ensure that the pipeline is working as it should be without bottlenecks or
performance issues. 

\section{Conclusion}

The article presents the \emph{ROS4HRI} framework. ROS4HRI
consists into two part: a model to flexibly represent humans for HRI
applications, and a transcription of this model into the ROS ecosystem.

Our human model has three important features: (1) it takes into account the
different requirements of different HRI applications by modularizing the model
into four parts (human body, human face, human voice and human `person') that
can be used independently or together; (2) it takes into account the
practicalities of social signal acquisition (like the importance of
re-identification) by introducing a system based on unique, transient IDs, that
enables a clean separation of concerns between (face, body, voice) detection on
one hand, and tracking and fusion on the other hand; (3) it does not make any
assumption regarding specific tools or package that could be used in an
implementation.

Our ROS implementation introduces a small set of new ROS messages (re-using
existing ones when sensible); set out a set of conventions regarding the
structure of HRI-related topics, tightly integrating the unique human IDs into the
naming scheme; introduce a kinematic model of the human that implements existing
ROS conventions, using dynamically generated URDF models to match the different
dimensions of each person, while leveraging existing ROS tools for eg
visualization.

Finally, the article introduces a ROS reference pipeline for HRI, as well as a
partial open-source implementation of the pipeline (including faces, bodies and
persons processing). The pipeline consists in new ROS wrappers around existing
software packages like OpenFace or OpenVINO, as well as entirely new nodes, like
the dynamic URDF generator or the `person' manager.

Together, these tree contributions (human model, ROS specification, and
reference implementation) will significantly contribute to close the `HRI gap'
in the ROS ecosystem. This article also aims at engaging the HRI community with
this specification effort, and, at the term of this process, we intend to submit
a new ROS REP to formally specify our model and conventions.

\bibliographystyle{acm}
\bibliography{BIB.bib}

\begin{thebibliography}{10}

\bibitem{badino2016integrating}
{\sc Badino, L., Canevari, C., Fadiga, L., and Metta, G.}
\newblock Integrating articulatory data in deep neural network-based acoustic
  modeling.
\newblock {\em Computer Speech \& Language 36\/} (2016), 173--195.

\bibitem{inproceedings}
{\sc Baltrusaitis, T., Robinson, P., and Morency, L.-P.}
\newblock Openface: An open source facial behavior analysis toolkit.
\newblock pp.~1--10.

\bibitem{bormann2013person}
{\sc Bormann, R., Zw{\"o}lfer, T., Fischer, J., Hampp, J., and H{\"a}gele, M.}
\newblock Person recognition for service robotics applications.
\newblock In {\em 2013 13th IEEE-RAS International Conference on Humanoid
  Robots (Humanoids)\/} (2013), IEEE, pp.~260--267.

\bibitem{burgoon2017social}
{\sc Burgoon, J.~K., Magnenat-Thalmann, N., Pantic, M., and Vinciarelli, A.}
\newblock {\em Social signal processing}.
\newblock Cambridge University Press, 2017.

\bibitem{ekman1992argument}
{\sc Ekman, P.}
\newblock An argument for basic emotions.
\newblock {\em Cognition \& emotion 6}, 3-4 (1992), 169--200.

\bibitem{ekman1978facs}
{\sc Ekman, P., and Friesen, W.}
\newblock Facial action coding system: A technique for the measurement of
  facial movement.

\bibitem{book}
{\sc Esposito, A., and Jain, L.}
\newblock {\em Toward Robotic Socially Believable Behaving Systems Volume II -
  “Modeling Social Signals”}, vol.~106.
\newblock 06 2016.

\bibitem{fong2006human}
{\sc Fong, T., Kunz, C., Hiatt, L.~M., and Bugajska, M.}
\newblock The human-robot interaction operating system.
\newblock In {\em Proceedings of the 1st ACM SIGCHI/SIGART conference on
  Human-robot interaction\/} (2006), pp.~41--48.

\bibitem{gaschler2012social}
{\sc Gaschler, A., Jentzsch, S., Giuliani, M., Huth, K., de~Ruiter, J., and
  Knoll, A.}
\newblock Social behavior recognition using body posture and head pose for
  human-robot interaction.
\newblock In {\em 2012 IEEE/RSJ International Conference on Intelligent Robots
  and Systems\/} (2012), IEEE, pp.~2128--2133.

\bibitem{10.5555/2906831.2906838}
{\sc Hebesberger, D., Dondrup, C., Koertner, T., Gisinger, C., and Pripfl, J.}
\newblock Lessons learned from the deployment of a long-term autonomous robot
  as companion in physical therapy for older adults with dementia: A mixed
  methods study.
\newblock In {\em The Eleventh ACM/IEEE International Conference on Human Robot
  Interaction\/} (2016), HRI '16, IEEE Press, p.~27–34.

\bibitem{king2009dlib}
{\sc King, D.~E.}
\newblock Dlib-ml: A machine learning toolkit.
\newblock {\em The Journal of Machine Learning Research 10\/} (2009),
  1755--1758.

\bibitem{kuo2009age}
{\sc Kuo, I.~H., Rabindran, J.~M., Broadbent, E., Lee, Y.~I., Kerse, N.,
  Stafford, R., and MacDonald, B.~A.}
\newblock Age and gender factors in user acceptance of healthcare robots.
\newblock In {\em RO-MAN 2009-The 18th IEEE International Symposium on Robot
  and Human Interactive Communication\/} (2009), IEEE, pp.~214--219.

\bibitem{lane-etal-2012-hritk}
{\sc Lane, I., Prasad, V., Sinha, G., Umuhoza, A., Luo, S., Chandrashekaran,
  A., and Raux, A.}
\newblock {HRI}tk: The human-robot interaction {T}ool{K}it rapid development of
  speech-centric interactive systems in {ROS}.
\newblock In {\em {NAACL}-{HLT} Workshop on Future directions and needs in the
  Spoken Dialog Community: Tools and Data ({SDCTD} 2012)\/} (Montr{\'e}al,
  Canada, June 2012), Association for Computational Linguistics, pp.~41--44.

\bibitem{lemaignan2016realtime}
{\sc Lemaignan, S., Garcia, F., Jacq, A., and Dillenbourg, P.}
\newblock From real-time attention assessment to “with-me-ness” in
  human-robot interaction.
\newblock In {\em Proceedings of the 2016 ACM/IEEE Human-Robot Interaction
  Conference\/} (2016).

\bibitem{lemaignan2017artificial}
{\sc Lemaignan, S., Warnier, M., Sisbot, E.~A., Clodic, A., and Alami, R.}
\newblock Artificial cognition for social human-robot interaction: An
  implementation.
\newblock {\em Artificial Intelligence\/} (2017).

\bibitem{li2013towards}
{\sc Li, R., Oskoei, M.~A., and Hu, H.}
\newblock Towards ros based multi-robot architecture for ambient assisted
  living.
\newblock In {\em 2013 IEEE International Conference on Systems, Man, and
  Cybernetics\/} (2013), IEEE, pp.~3458--3463.

\bibitem{michael2015domain}
{\sc Michael, J., and D’Ausilio, A.}
\newblock Domain-specific and domain-general processes in social perception--a
  complementary approach.
\newblock {\em Consciousness and cognition 36\/} (2015), 434--437.

\bibitem{quigley2009ros}
{\sc Quigley, M., Conley, K., Gerkey, B., Faust, J., Foote, T., Leibs, J.,
  Wheeler, R., and Ng, A.~Y.}
\newblock Ros: an open-source robot operating system.
\newblock In {\em ICRA workshop on open source software\/} (2009), vol.~3,
  Kobe, Japan, p.~5.

\bibitem{schuller2009interspeech}
{\sc Schuller, B., Steidl, S., and Batliner, A.}
\newblock The interspeech 2009 emotion challenge.
\newblock In {\em Tenth Annual Conference of the International Speech
  Communication Association\/} (2009).

\bibitem{Twomey_Morse_Cangelosi_Horst_2016}
{\sc Twomey, K.~E., Morse, A.~F., Cangelosi, A., and Horst, J.~S.}
\newblock Children’s referent selection and word learning.
\newblock {\em Interaction Studies 17}, 1 (Sep 2016), 101–127.

\bibitem{wagner2013social}
{\sc Wagner, J., Lingenfelser, F., Baur, T., Damian, I., Kistler, F., and
  Andr{\'e}, E.}
\newblock The social signal interpretation (ssi) framework: multimodal signal
  processing and recognition in real-time.
\newblock In {\em Proceedings of the 21st ACM international conference on
  Multimedia\/} (2013), pp.~831--834.

\bibitem{zhang2015ros}
{\sc Zhang, Y., and Xu, S.~C.}
\newblock Ros based voice-control navigation of intelligent wheelchair.
\newblock In {\em Applied Mechanics and Materials\/} (2015), vol.~733, Trans
  Tech Publ, pp.~740--744.

\end{thebibliography}

\end{document}